\documentclass{article}
\usepackage{ijcai24}

\usepackage{times}
\usepackage{soul}
\usepackage{url}
\usepackage[hidelinks]{hyperref}
\usepackage[utf8]{inputenc}
\usepackage[small]{caption}
\usepackage{graphicx}
\usepackage{amsmath}
\usepackage{amsthm}
\usepackage{booktabs}
\usepackage{algorithm}
\usepackage{algorithmic}
\usepackage[switch]{lineno}
\usepackage{multirow}
\usepackage{hhline}
\usepackage{pifont}
\usepackage{array}

\title{Unified Unsupervised Salient Object Detection via Knowledge Transfer}

\author{
Yao Yuan
\and
Wutao Liu\and
Pan Gao\footnote{Corresponding author} \and
Qun Dai\And
Jie Qin\footnotemark[1]
\affiliations
Nanjing University of Aeronautics and Astronautics
\emails
\{ayews233, wutaoliu, pan.gao, daiqun\}@nuaa.edu.cn,
qinjiebuaa@gmail.com
}

\begin{document}
\maketitle

\begin{abstract}
Recently, unsupervised salient object detection (USOD) has gained increasing attention due to its annotation-free nature. 
However, current methods mainly focus on specific tasks such as RGB and RGB-D, neglecting the potential for task migration. 
In this paper, we propose a unified USOD framework for generic USOD tasks. 
Firstly, we propose a Progressive Curriculum Learning-based Saliency Distilling (PCL-SD) mechanism to extract saliency cues from a pre-trained deep network. 
This mechanism starts with easy samples and progressively moves towards harder ones, to avoid initial interference caused by hard samples. 
Afterwards, the obtained saliency cues are utilized to train a saliency detector, and we employ a Self-rectify Pseudo-label Refinement (SPR) mechanism to improve the quality of pseudo-labels. 
Finally, an adapter-tuning method is devised to transfer the acquired saliency knowledge, leveraging shared knowledge to attain superior transferring performance on the target tasks. 
Extensive experiments on five representative SOD tasks confirm the effectiveness and feasibility of our proposed method. Code and supplement materials are available at https://github.com/I2-Multimedia-Lab/A2S-v3. 
\end{abstract}

\section{Introduction}

Salient object detection (SOD) aims to identify the most visually significant objects in images. Supervised SOD methods have achieved excellent results, but due to their heavy reliance on pixel-level annotations for salient objects, unsupervised SOD (USOD) has been gaining increasing attention. USOD not only eliminates the need for annotated data but also exhibits strong generalization performance when applied to other tasks~\cite{9123596,Wu_2021_ICCV}. 

Traditional SOD methods rely heavily on hand-crafted features, such as color and contrast, for saliency extraction. Although these methods prove effective for simple scenes, they encounter difficulties in complex scenes due to the absence of high-level semantic information. Existing deep learning-based USOD methods ~\cite{nguyen2019deepusps,zhang2018deep} mostly utilize the predictions generated by traditional SOD methods as saliency cues and incorporate semantic information to generate refined saliency predictions. 
Recently, based on the observation that CNNs pre-trained on large-scale data usually produce high activations on some primary objects, 
A2S~\cite{zhou2023a2s1} have developed a method to distill saliency from the activation maps of deep networks and generate high-quality pseudo-labels. 
However, we found that during the initial training phase, the presence of hard samples in complex scenes or along object boundaries results in the accumulation of irreparable errors. 
\begin{figure}
	\centering 
	\includegraphics[scale=0.58]{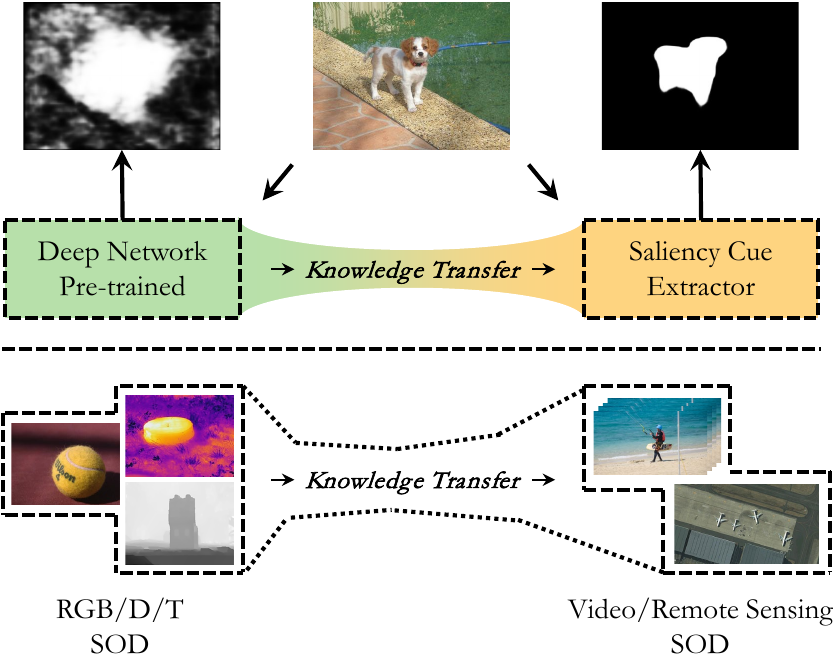}
 	\vspace{-2mm}
	\caption{The proposed framework includes two types of knowledge transfer: (1) From pre-trained deep network to saliency cue extractor; (2) From Natural Still Image (NSI) SOD to non-NSI SOD. }
	\label{fig:Knowledge}
	\vspace{-4mm}
\end{figure}

Unsupervised SOD is generally considered to 
exhibit strong generalization and transferability due to its annotation-free nature. 
However, prevailing USOD methodologies predominantly focus on the Natural Still Image (NSl) domain, 
exemplified by RGB, RGB-D, and RGB-T. 
Consequently, USOD on non-NSI domain, encompassing video SOD and Remote Sensing Image (RSI) SOD, remains largely unexplored, presenting a notable research gap in the field. 
We believe that different SOD tasks share common knowledge, 
and exploiting this shared knowledge can benefit transfer performance. 
On the other hand, compared to NSI SOD, the available datasets for video SOD or RSI SOD are relatively small and burdensome to obtain. 
As a result, training models from scratch on these tasks to obtain satisfactory performance is currently deemed impractical. 
Therefore, we advocate for the investigation of a more universally applicable unsupervised saliency knowledge transfer method. 


To address the aforementioned issues, we design a unified framework for generic unsupervised SOD tasks. 
Firstly, we propose the Progressive Curriculum Learning-based Saliency Distilling (PCL-SD) mechanism to guide the extraction of saliency cues. 
At the early stages of training, we only extract preliminary saliency cues from easy samples. 
As the training progresses, we progressively incorporate hard samples to mine deeper saliency knowledge. 
The employment of PCL-SD effectively mitigates the initial accumulation of errors, leading to a more stable and robust training process. 
Next, we utilize the obtained saliency cues to train a saliency detector and design 
a Self-rectify Pseudo-label Refinement (SPR) mechanism to improve the quality of pseudo-labels. 
On one hand, the proposed SPR employs the saliency predictions of the model during training to rectify incorrect predictions within the pseudo-labels. 
On the other hand, it incorporates the prior knowledge of the input image to prevent the model from becoming complacent. 
The SPR mechanism demonstrates a strong capability in self-supervised learning, resulting in improved pseudo-label quality. 
Finally, we devise an adapter-tuning method to transfer the acquired saliency knowledge to non-NSI SOD tasks, such as video SOD and RSI SOD. 
Specifically, we selectively fine-tune the deep features, ensuring effective adaptation of the model to the target task while mitigating the risk of model degradation.


Our main contributions can be summarized as follows: 
\begin{itemize}
	\item We propose the Progressive Curriculum Learning-based Saliency Distilling (PCL-SD) mechanism to extract saliency cues from easy samples to hard ones. 
	
	\item We design the Self-rectify Pseudo-label Refinement (SPR) mechanism to gradually improve the quality of pseudo-labels during the training process. 
	
	\item We devise an adapter-tuning method to transfer saliency knowledge from NSI SOD to non-NSI SOD tasks, achieving impressive transfer performance. 

\end{itemize}
Note that we are the first to consider knowledge transfer from NSI domain to non-NSI domain, and develop a unified framework for generic USOD tasks. 
Experiments on RGB, RGB-D, RGB-T, video SOD and RSI SOD benchmarks confirm the state-of-the-art USOD performance of our method. 

\section{Related Works}
%

\subsection{Unsupervised Salient Object Detection}
Traditional SOD methods rely on hand-crafted features to extract saliency cues. For instance, ~\cite{perazzi2012saliency} estimates saliency by evaluating the contrast in uniqueness and spatial distribution within the image. ~\cite{jiang2011automatic} employ a combination of bottom-up salient stimuli and object-level shape prior to segment salient objects. Although these approaches perform well in simple scenes, they face challenges in handling complex scenes due to the lack of high-level semantic information. 

Existing deep learning-based methods for USOD typically involve two stages. In the first stage, pseudo-labels are obtained, while in the second stage, a network is trained using these pseudo-labels. 
For instance, ~\cite{zhang2017supervision} fuses multiple noisy saliency cues to generate supervisory signals for training the deep salient object detector. In \cite{nguyen2019deepusps}, a set of refinement networks are initially trained to enhance the quality of these saliency cues, and the refined pseudo-labels are subsequently utilized to train a saliency detector. A more recent approach, A2S~\cite{zhou2023a2s1}, proposes a method to distill saliency from the activation maps of deep networks, achieving high-quality pseudo-labels. 

\subsection{Knowledge Transfer in SOD}
Knowledge transfer involves applying models or features trained in one task or domain to another related task or domain. 
A typical example is fine-tuning a deep network that was pre-trained on large-scale data for a specific target task. 
However, the exploration of knowledge transfer across different SOD tasks remains insufficient. 
Among the limited studies, \cite{FU2022142} addresses the RGB-D SOD task as a few-shot learning problem 
and enhances performance by incorporating knowledge from RGB SOD. 
\cite{zhou2023texture} employs data from multiple SOD tasks to train a generalized saliency detector. 
Nevertheless, when extending to generic SOD tasks, 
the inherent gap between various SOD tasks can impede effective model training. 
Consequently, it becomes crucial to devise a knowledge transfer approach that is rooted in shared knowledge. 
\section{Proposed Method}
\begin{figure*}
	\centering 
	\includegraphics[scale=0.87]{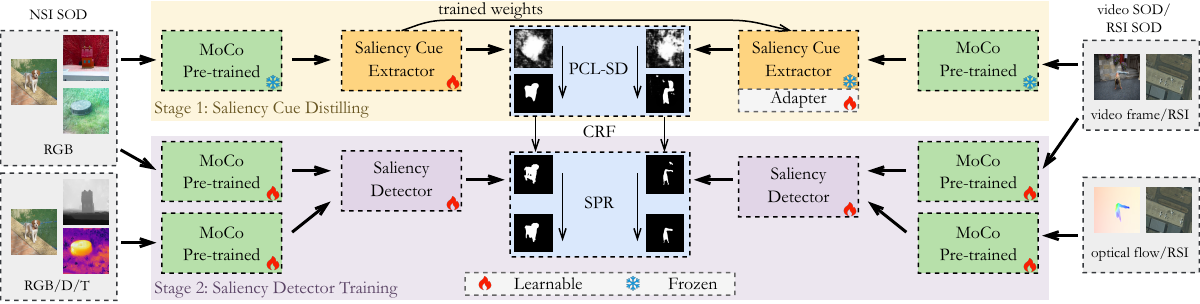}
	\caption{Overview of the proposed method. 
	The left side represents the training process on NSI SOD, while the right side shows the training process of transferring to non-NSI SOD tasks. }
	\label{fig:model}
	\vspace{-2mm}
\end{figure*}

Figure \ref{fig:model} illustrates the proposed two-stage framework. 
In stage 1, we train a saliency cue extractor (SCE) to transfer saliency knowledge from a pre-trained deep network. 
The proposed Progressive Curriculum Learning-based Saliency Distilling is employed to 
mitigate the initial accumulation of errors in training and ensure the stability and robustness of the training process. 
In stage 2, we utilize the obtained saliency cues as initial pseudo-labels to train a saliency detector (SD). 
CRF~\cite{krahenbuhl2011efficient} is adopted to enhance the initial pseudo-labels, and we employ the proposed Self-rectify Pseudo-label Refinement mechanism to improve pseudo-labels quality during the training process gradually.  

Initially, we train our base model on Natural Still Image (NSI) SOD and subsequently transfer the model to non-NSI SOD tasks. 
Throughout the training of the base model, we combine all the NSI data for training. However, during the transfer process, 
we only employ task-specific data for training. 
For example, when migrating to video SOD, we solely utilize video frames and optical flow as input. 
The transfer process also follows a two-stage training approach, while we applied the proposed fine-tuning method to optimize the SCE instead of training it from scratch. 
Besides, ResNet-50~\cite{ResNet} pre-trained by MoCo-v2~\cite{chen2020improved}, A2S~\cite{zhou2023a2s1} and MIDD~\cite{9454273} are employed as the pre-trained deep network, SCE, and SD, respectively. 
A more detailed description and explanation can be found in supplementary materials. 
\subsection{Progressive Curriculum Learning-based Saliency Distilling}
\begin{figure}
	\centering 
	\includegraphics[scale=0.84]{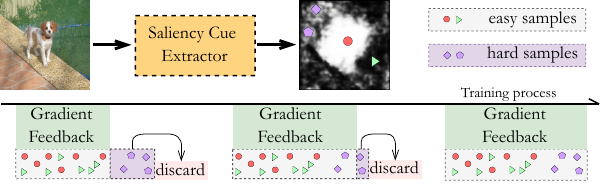}
	\caption{Illustration of the proposed PCL-SD. Hard samples are progressively incorporated as the training progresses. }
	\label{fig:pcl-sd}
	\vspace{-2mm}
\end{figure}
The problem of obtaining saliency cues or extracting salient information from scratch 
has always been a challenge for unsupervised salient object detection methods. 
Earlier deep learning-based methods~\cite{zhang2017supervision} relied on noisy saliency cues 
generated by traditional SOD methods, 
while approaches like A2S~\cite{zhou2023a2s1} employ the activation maps 
produced by a pre-trained network as saliency cues. 
This method effectively extracts the saliency information embedded in the pre-trained network. 
However, at the early stages of training, 
hard samples in complex scenes may corrupt the fragile saliency patterns in the network, 
leading to irreparable accumulation errors and the risk of pattern collapse. 
To address this issue, we introduce the concept of curriculum learning 
into saliency distilling and propose 
Progressive Curriculum Learning-based Saliency Distilling (PCL-SD). 
As can be seen in Figure \ref{fig:pcl-sd}, the proposed PCL-SD rigidly excludes hard samples 
at the early stages of training and gradually incorporates them as training progresses. 
As a result, the model progressively extracts saliency knowledge from easy to hard samples, 
and the entire training process becomes more robust and stable. 

The process of saliency distilling can be formulated as:
\begin{equation}
	\small
	\begin{split}
		&	\mathcal{L}_{sal} = 0.5-\frac{1}{N}\sum_{i}^{N}\|S(i)-0.5|
	\end{split}
\end{equation}
Here, $N$ represents the number of pixels, and $S(i)$ denotes pixel $i$ in the saliency prediction $S$ output by the saliency cue extractor (SCE). 
To be intuitively described, $\mathcal{L}_{sal}$ pulls the predicted values of each pixel towards either 0 or 1. 
However, during the early stages of training, $\mathcal{L}_{sal}$ may pull hard samples with values close to 0.5 in the wrong direction, which we refer to as the problem of error accumulation. 
The proposed PCL-SD strategy focuses on two essential aspects: 
(1) how to define hard samples, and (2) how to gradually incorporate them. 
Firstly, the determination of a pixel in the saliency prediction $S$ as a hard sample is based on its prediction value. 
Specifically, a pixel $S(i)$ is classified as a hard sample if 
\begin{equation}
	\small
	\begin{split}
		&	|S(i)-0.5| < p. 
	\end{split}
\end{equation}
Here, $p$ is the threshold for dividing hard samples, 
with a larger $p$ indicating more hard samples are divided. 
Secondly, the value of $p$ is initially set as 0.2 and progressively decreased 
during the training process until all samples are included. 
This decrease is governed by the formula: 
\begin{equation}
	\small
	\begin{split}
		&	p = Max(0, 0.2-0.6\times E_c/E_t), 
	\end{split}
\end{equation}
where $E_c$ and $E_t$ denote current epoch and total epoch, respectively. 
Finally, we define PCL-SD as: 
\begin{equation}
	\small
	\begin{split}
		&			M(i) = \left \{ 
				\begin{matrix} 
					&	0 & if\ |S(i)-0.5| < p, \\ 
					&	1 & otherwise, 
				\end{matrix}\right.\\
		&	\mathcal{L}_{pcl-sd} = 0.5-\frac{1}{N}\sum_{i}^{N}|M(i)\odot S(i)-0.5|
	\end{split}
\end{equation}
where $\odot$ denotes the Hadamard product for matrices. 
\subsection{Self-rectify Pseudo-label Refinement}

\begin{figure}
	\centering 
	\includegraphics[scale=0.84]{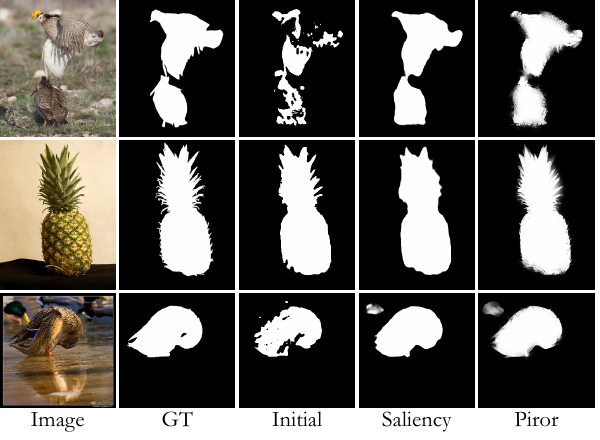}
	\vspace{-6mm}
	\caption{The comparison between initial pseudo-label, saliency prediction, and prior rectification. }
	\label{fig:spr}
	\vspace{-2mm}
\end{figure}

Obtaining high-quality pseudo-labels is crucial for training a saliency detector (SD). 
On the other hand, as shown in Figure \ref{fig:spr}, the saliency prediction $S$ output by SD can partially rectify errors within the pseudo-labels. 
We define saliency prediction as posterior rectification: $R_{\text{post}} = S $. 
However, while this posterior rectification can rectify errors in initial pseudo-labels, 
it also introduces the risk of the model becoming overly confident and stagnant. 
To overcome this, we introduce prior information from the input image to optimize saliency prediction, 
in order to avoid the model falling into a self-complacent trap. 

Previous approaches primarily rely on CRF for prior rectification, which entails significant computational costs. 
Inspired by ~\cite{ru2022learning}, we employ a real-time pixel refiner to provide efficient prior rectification based on the input image. 
To start, we define the feature distance $d_f^{i,j}$ and position distance $d_p^{i,j}$ between pixels as follows: 
\begin{equation}
	\small
	\begin{split}
		&	d_f^{i,j} = -\frac{\|I(i)-I(j)\|}{\omega_1 \sigma_f}, d_p^{ij} = -\frac{\|P(i)-P(j)\|}{\omega_2 \sigma_p}
	\end{split}
\end{equation}
Here, $I$ and $P$ represent the input image and position information, 
while $\sigma_f$ and $\sigma_p$ denote the standard deviation of feature values and position differences, respectively. 
The parameters $\omega_1$ and $\omega_2$ control the smoothness. The refiner $R(\cdot)$ is then defined as: 
\begin{equation}
	\small
	\begin{split}
		&	R(I) = \sum_{j \in \mathcal{N}(i)}^{}(\frac{exp(d_f^{ij})}{\sum\limits_{k \in \mathcal{N}(i)}exp(d_f^{ik})}+\omega_3\frac{exp(d_p^{ij})}{\sum\limits_{k \in \mathcal{N}(i)}exp(d_p^{ik})})
	\end{split}
\end{equation} 
Here, $\mathcal N(\cdot)$ represents the set of neighboring pixels in an 8-way manner. Finally, the prior rectification can be defined as: 
\begin{equation}
\small
\begin{split}
& R_{\text{pri}} = R(I) \odot S
\end{split}
\end{equation}
where $\odot$ denotes the Hadamard product for matrices. At last, the refined pseudo-label is defined as:
\begin{equation}
\small
\begin{split}
& G_{\text{ref}} = \lambda_1 R_{\text{pri}} + \lambda_2 R_{\text{post}} + \lambda_3 G_{\text{pre}}
\end{split}
\end{equation}
Here, $G_{\text{ref}}$ refers to the pseudo-labels after refinement, and $G_{\text{pre}}$ refers to the previous pseudo-labels. 
The introduction of $G_{\text{ref}}$ aims to improve the stability of the refinement process. 
$\lambda_1$, $\lambda_2$, $\lambda_3$ are empirically assigned as 0.2, 0.6, and 0.2, respectively. 
As shown in Figure \ref{fig:spr}, prior rectification has effectively compensated for the considerable loss of local details. 
The proposed SPR mechanism combines prior and posterior rectification, 
gradually improving the quality of pseudo labels during the training process, 
demonstrating strong self-supervised performance. 

\subsection{Knowledge Transfer to non-NSI SOD tasks}

\begin{figure}
	\centering 
	\includegraphics[scale=0.45]{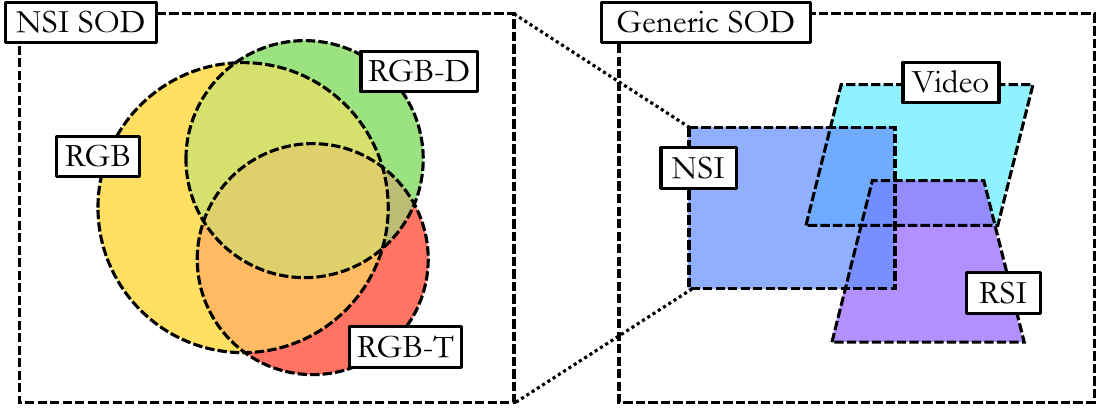}
	\caption{The relevance between different SOD tasks. The overlaps can be seen as shared common knowledge. }
	\label{fig:domain}
	\vspace{-4mm}
\end{figure}

We investigate the transferability of the proposed method to video SOD and Remote Sensing Image (RSI) SOD. 
Figure \ref{fig:domain} illustrates the varying degrees of relevance between different SOD tasks. 
The tasks within NSI SOD benefit from a greater amount of shared knowledge, 
allowing for the joint training of multiple tasks to achieve a better generalization performance. 
However, as we broaden our focus to generic SOD tasks, the inherent gap between tasks becomes the primary influencing factor. 
Joint training becomes more challenging and poses risks of model degradation. 
More discussions on this topic can be found in supplementary materials. 

We posit that identifying an appropriate fine-tuning method can effectively address the issue of model degradation. 
Inspired by recent studies on Adapter-tuning~\cite{houlsby2019parameter}, 
we design a simple but effective fine-tuning method for knowledge transfer from NSI SOD to non-NSI SOD tasks. 
Specifically, for end-to-end tasks in SOD, the prevailing methods and models employ the U-net~\cite{Unet} structure 
and utilize multi-scale feature aggregation to achieve accurate saliency predictions. 
We contend that shallow features primarily contribute to local details and possess a degree of cross-task generality, 
while deep features play a pivotal role in salient object localization and exhibit task-specific characteristics. 
Hence, we suppose that fine-tuning solely the network layers or modules responsible for deep feature handling 
allows the model to adapt to the target task while circumventing degradation. 
Technically, we define the deep feature handling process in the model as 
\begin{equation}
\small
\begin{split}
& \hat{F} = T(F)
\end{split}
\end{equation}
Here, $F$ represents the deep features extracted by the backbone, $\hat{F}$ denotes the processed features, and $T$ signifies the network layer or module performing the processing. In specific end-to-end SOD models, $T$ can comprise a convolutional layer that modifies the number of feature channels or a network module that enhances the features. Our adapter-tuning approach can be defined as: 
\begin{equation}
\small
\begin{split}
& \hat{F} = T(F) + T_a(F)
\end{split}
\end{equation}
In this equation, $T_a$ refers to the adapter, which possesses a structure consistent with $T$. Following the processing, $T_a$ is connected to the original network through a residual connection. During fine-tuning, we exclusively optimize the weights of $T_a$ while keeping the remaining weights of the model frozen. 
The detailed description can be found in supplementary materials. 
It is worth mentioning that this fine-tuning method is universal for any kind of SOD method or task. 
\subsection{Supervision Strategy}
We initially train the saliency cue extractor (SCE) in the first stage, followed by training the saliency detector (SD) in the second stage. In the training of the first stage, we also incorporate Boundary-aware Texture Matching (BTM)~\cite{zhou2023texture} to introduce extra structural cues, and is formulated as: 
\begin{equation}
\small
\begin{split}
& \mathcal{L}_{btm} = \frac{\sum_i b_iT_i^s·(T_i^a)^T}{\sum_i b_i}. 
\end{split}
\end{equation}
Here, $T_i^s$ represents the saliency texture vector, $T_i^a$ denotes the input image texture vector, 
and $b_i$ represents the binary boundary mask of the saliency prediction. 
Moreover, a structural consistency loss is employed to achieve transformation-invariant predictions, and is formulated as: 
\begin{equation}
\small
\begin{split}
& \mathcal{L}_{sc} = \sum_i^N\|S(i)-\hat{S}(i)\|. 
\end{split}
\end{equation}
Here, $\hat{S}$ denotes saliency prediction after transformation. 
To ensure training stability, only random scaling is adopted. 
The total loss for training SCE can be defined as: 
\begin{equation}
\small
\begin{split}
& \mathcal{L}_{sce} = \mathcal{L}_{pcl-sd}+\gamma \mathcal{L}_{btm}+\mathcal{L}_{sc}. 
\end{split}
\end{equation}
$\gamma$ is empirically assigned as 0.05. 

We train the saliency detector (SD) with IoU loss, which is defined as:
\begin{equation}
	\small
	\begin{split}
	\mathcal{L}_{IoU}=1-\frac{\sum_{i=1}^{N}(S(i)G(i))}{\sum_{i=1}^{N}(S(i)+G(i)-S(i)G(i))},
	\end{split}
\end{equation}
$G$ refers to the pseudo-labels, and the total loss for training SD can be defined as: 
\begin{equation}
\small
\begin{split}
\mathcal{L}_{sd} = \mathcal{L}_{IoU}+\mathcal{L}_{sc}. 
\end{split}
\end{equation}
\section{Experiments}
\subsection{Implementation Details}
\subsubsection{Training Settings}
The batch size is set to 8 and input images are resized to 320$\times$320. 
Horizontal flipping is employed as our data augmentation. 
We train the saliency cue extractor for 20 epochs using the SGD optimizer with an initial learning rate of 0.1, which is decayed linearly. 
We train the saliency detector for 10 epochs using the SGD optimizer with a learning rate of 0.005. 
All experiments were implemented on a single RTX 3090 GPU. 
\subsubsection{Datasets}
We follow the prevalent settings of SOD and relevant tasks. Here are some details about the datasets we used.
\textbf{(1) RGB SOD:} We use the training subsets of DUTS~\cite{DUTS} to train our method. 
ECSSD~\cite{ECSSD}, PASCALS~\cite{PASCAL-S}, HKU-IS~\cite{HKU-IS}, DUTS-TE~\cite{DUTS} and DUT-O~\cite{DUT-O} are employed for evaluation.
\textbf{(2) RGB-D SOD:} We choose 2185 samples from the training subsets of NLPR~\cite{peng2014rgbd} and NJUD~\cite{ju2014depth} as the training set. 
RGBD135~\cite{cheng2014depth}, SIP~\cite{fan2020rethinking} and the testing subsets of NJUD and NLPR are employed for evaluation. 
\textbf{(3) RGB-T SOD:}
2500 images in VT5000~\cite{tu2022rgbt} are for training, while VT1000~\cite{tu2019rgb}, VT821~\cite{wang2018rgb} and the rest 2500 images in VT5000 are for testing. 
\textbf{(4) Video SOD:} We choose the training splits of DAVIS~\cite{perazzi2016benchmark} and DAVSOD~\cite{fan2019shifting} to train our method. 
SegV2~\cite{li2013video}, FBMS~\cite{brox2010object} and the testing splits of DAVIS and DAVSOD are employed for evaluation. 
\textbf{(5) Remote Sensing Image SOD:} We choose the training splits of ORSSD~\cite{li2019nested} and EORSSD~\cite{zhang2020dense} to train our method. 
The testing splits of ORSSD and EORSSD are employed for evaluation. 

\subsubsection{Metrics}
We employ three metrics to evaluate our model and the existing state-of-the-art methods, 
including Mean Absolute Error $M$, average F-measure ($F_\beta$)~\cite{Achanta2009FrequencytunedSR} and E-measure ($E_\xi$)~\cite{ijcai2018p97}. 
Specifically, ($M$) measures the average pixel-wise difference between the prediction $P$ and the ground truth $G$, 
and is calculated as $M = \frac{1}{N}\sum_{i=1}^{N}\vert P(i)-G(i)\vert$. 
$F_\beta$ considers both precision and recall values of the prediction map, 
and can be computed as $F_\beta=\frac{(1+\beta^2)\times Precision\times Recall}{\beta^2\times Precision+Recall}$, with $\beta^2$ set to 0.3. 
$E_\xi$ takes into account the local pixel values along with the image-level mean value, 
and is defined as $E_\xi=\frac{1}{N}\sum_{i=1}^{N}\phi_\xi(i,j)$, 
where $\phi_\xi$ represents the enhanced alignment matrix. 



\subsection{Comparisons With State-of-the-Art}
We report the performance of our method on five representative SOD tasks, and more qualitative results are provided in supplementary materials.
\begin{figure*}
	\centering 
	\includegraphics[scale=0.85]{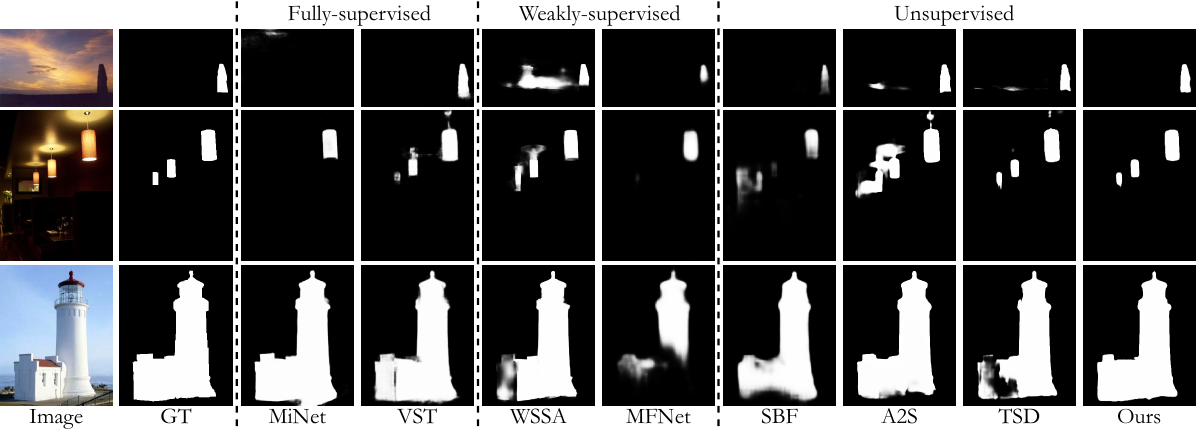}
 	\vspace{-2mm}
	\caption{Visual comparison between the proposed method and  other state-of-the-art SOD methods on RGB SOD datasets. }
	\label{fig:visual}
	\vspace{-2mm}
\end{figure*}
\subsubsection{Results on RGB SOD}
Table \ref{tab:RGBSOD} presents a quantitative comparison between the proposed method and recent fully-supervised, weakly-supervised, and unsupervised methods. 
The fully-supervised methods include MINet~\cite{MiNet} and VST~\cite{VST}, 
the weakly-supervised methods contain WSSA~\cite{zhang2020weakly} and MFNet~\cite{Piao_2021_ICCV}, 
and the unsupervised methods comprise SBF~\cite{zhang2017supervision}, EDNS~\cite{zhang2020learning}, DCFD~\cite{lin2022causal}, 
SelfMask~\cite{shin2022unsupervised}, TSD~\cite{zhou2023texture} and STC~\cite{song2023towards}. 
Our results are presented under different settings: 
(1) Training our method using task-specified data, denoted as ``Ours$_{t.s.}$", for a fair comparison; 
(2) Training our method using Nature Still Image (NSI) data, including RGB, RGB-D, and RGB-T datasets, referred to as ``Ours". 

The results presented in Table \ref{tab:RGBSOD} clearly indicate that the proposed method outperforms existing USOD methods, 
leading to significant improvements in performance. 
Additionally, our unsupervised approach demonstrates competitive performance in comparison to recent weakly-supervised and fully-supervised methods. 
Notably, our method, referred to as ``Ours", exhibits a slight superiority over ``Ours$_{t.s.}$". 
We suppose that this advantage stems from the utilization of a more extensive training dataset, 
which enhances the model's generalization ability and leads to improved performance when applied to unseen images. 

A qualitative comparison is presented in Figure \ref{fig:visual}. As can be seen, our method has achieved more accurate and complete saliency prediction. 
Moreover, our approach exhibits remarkable performance in dealing with multiple objects (row 2). 
\begin{table*}
	\centering
	\small
	\setlength\tabcolsep{1mm}
	\begin{tabular}{l|c|c|ccc|ccc|ccc|ccc|ccc}%
		\hline 
		\multicolumn{3}{c|}{dataset} & \multicolumn{3}{c|}{DUT-O} & \multicolumn{3}{c|}{DUTS-TE} & \multicolumn{3}{c|}{ECSSD} & \multicolumn{3}{c|}{HKU-IS} & \multicolumn{3}{c}{PASCAL-S}\\
		\hline 
		Method & Year & Sup. & $M\downarrow\:$ & $F_\beta\uparrow\:$ & $E_\xi\uparrow\:$ & $M\downarrow\:$ & $F_\beta\uparrow\:$ & $E_\xi\uparrow\:$ & $M\downarrow\:$ & $F_\beta\uparrow\:$ & $E_\xi\uparrow\:$ & $M\downarrow\:$ & $F_\beta\uparrow\:$ & $E_\xi\uparrow\:$ & $M\downarrow\:$ & $F_\beta\uparrow\:$ & $E_\xi\uparrow\:$ \\
		\hline
		MINet & 2020 & F & \textbf{.055} & \textbf{.756} & \textbf{.873} & \textbf{.037} & \textbf{.828} & \textbf{.917} & \textbf{.033} & \textbf{.924} & .953 & \textbf{.028} & \textbf{.908} & \textbf{.961} & .064 & \textbf{.842} & .899 \\
		VST & 2021 & F & .058 & \textbf{.756} & .872 & \textbf{.037} & .818 & .916 & \textbf{.033} & .92 & \textbf{.957} & .029 & .9 & .96 & \textbf{.061} & .829 & \textbf{.902} \\
		\hline
		WSSA & 2020 & W & \textbf{.068} & \textbf{.703} & \textbf{.845} & \textbf{.062} & \textbf{.742} & \textbf{.869} & \textbf{.047} & \textbf{.860} & \textbf{.932} & \textbf{.059} & \textbf{.870} & \textbf{.917} & \textbf{.096} & \textbf{.785} & \textbf{.855} \\
		MFNet & 2021 & W & .098 & .621 & .784 & .079 & .693 & .832 & .058 & .839 & .919 & .084 & .844 & .889 & .115 & .756 & .824 \\
		\hline
		EDNS & 2020 & U & .076 & .682 & .821 & .065 & .735 & .847 & .068 & .872 & .906 & .046 & .874 & .933 & .097 & .801 & .846 \\
		SelfMask & 2022 & U & .078 & .668 & .815 & .063 & .714 & .848 & .058 & .856 & .920 & .053 & .819 & .915 & .087 & .774 & .856 \\
		DCFD & 2022 & U & .070 & .710 & .837 & .064 & .764 & .855 & .059 & .888 & .915 & .042 & .889 & .935 & .090 & .795 & .860 \\
		TSD & 2023 & U & \textbf{.061} & .745 & .863 & .047 & .810 & .901 & .044 & .916 & .938 & .037 & .902 & .947 & .074 & .830 & .882 \\
		STC & 2023 & U & .068 & .753 & .852 & .052 & .809 & .891 & .050 & .903 & .935 & .041 & .891 & .942 & .076 & .827 & .881 \\
		Ours$_{t.s.}$ & - & U & .063 & .749 & .864 & \textbf{.046} & .814 & \textbf{.906} & \textbf{.038} & .922 & .95 & .034 & .905 & .953 & \textbf{.068} & .841 & .898 \\
		Ours & - & U & {.062} & \textbf{.759} & \textbf{.868} & .047 & \textbf{.816} & \textbf{.906} & \textbf{.038} & \textbf{.923} & \textbf{.951} & \textbf{.033} & \textbf{.908} & \textbf{.954} & .069 & \textbf{.844} & \textbf{.899} \\
		\hline 
	\end{tabular}
	\vspace{-1mm}
	\caption{Quantitative comparison on RGB SOD benchmarks. “Sup.” indicates the supervised signals used to train SOD methods. ``F", ``W" and ``U" mean fully-supervised, weakly-supervised and unsupervised, respectively. The best results are shown in \textbf{bold}.}
	\label{tab:RGBSOD}
	\vspace{-2mm}
\end{table*}
\begin{table*}
	\centering
	\small
	\setlength\tabcolsep{1.5mm}

	\begin{tabular}{l|c|c|ccc|ccc|ccc|ccc}%
		\hline 
		\multicolumn{3}{c|}{dataset} & \multicolumn{3}{c|}{RGBD-135} & \multicolumn{3}{c|}{NJUD} & \multicolumn{3}{c|}{NLPR} & \multicolumn{3}{c}{SIP} \\
		\hline 
		Method & Year & Sup. & $M\downarrow\:$ & $F_\beta\uparrow\:$ & $E_\xi\uparrow\:$ & $M\downarrow\:$ & $F_\beta\uparrow\:$ & $E_\xi\uparrow\:$ & $M\downarrow\:$ & $F_\beta\uparrow\:$ & $E_\xi\uparrow\:$ & $M\downarrow\:$ & $F_\beta\uparrow\:$ & $E_\xi\uparrow\:$ \\
		\hline
		VST & 2021 & F & \textbf{.017} & \textbf{.917} & \textbf{.979} & .034 & .899 & .943 & .023 & .886 & .956 & \textbf{.04} & \textbf{.895} & \textbf{.941} \\
		CCFE & 2022 & F & .020 & .911 & .964 & \textbf{.032} & \textbf{.914} & \textbf{.953} & \textbf{.021} & \textbf{.907} & \textbf{.962} & .047 & .889 & .923 \\
		\hline
		DSU & 2022 & U & .061 & .767 & .895 & .135 & .719 & .797 & .065 & .745 & .879 & .156 & .619 & .774 \\
		TSD & 2023 & U & .029 & 877 & \textbf{.946} & .060 & .862 & .908 & .034 & .852 & .931 & .051 & .873 & .925 \\
		Ours$_{t.s.}$ & - & U & .027 & .882 & {.945} & .053 & .862 & .915 & .034 & .853 & .935 & .042 & .876 & \textbf{.935} \\
		Ours & - & U & \textbf{.025} & \textbf{.888} & .94 & \textbf{.049} & \textbf{.876} & \textbf{.923} & \textbf{.028} & \textbf{.871} & \textbf{.945} & \textbf{.04} & \textbf{.879} & .931 \\
		\hline 
	\end{tabular}
	\vspace{-1mm}
	\caption{Quantitative comparison on RGB-D SOD benchmarks. }
	\label{tab:RGBDSOD}
	\vspace{-2mm}
\end{table*}
\begin{table*}
	\centering
	\small
	\setlength\tabcolsep{2.5mm}
	\begin{tabular}{l|c|c|ccc|ccc|ccc}%
		\hline 
		\multicolumn{3}{c|}{dataset} & \multicolumn{3}{c|}{VT5000} & \multicolumn{3}{c|}{VT1000} & \multicolumn{3}{c}{VT821} \\
		\hline 
		Method & Year & Sup. & $M\downarrow\:$ & $F_\beta\uparrow\:$ & $E_\xi\uparrow\:$ & $M\downarrow\:$ & $F_\beta\uparrow\:$ & $E_\xi\uparrow\:$ & $M\downarrow\:$ & $F_\beta\uparrow\:$ & $E_\xi\uparrow\:$  \\
		\hline
		MIDD & 2021 & F & .043 & .801 & .899 & .027 & .882 & .942 & .045 & .805 & .898 \\
		CCFE & 2022 & F & \textbf{.030} & \textbf{.859} & \textbf{.937} & \textbf{.018} & \textbf{.906} & \textbf{.963} & \textbf{.027} & \textbf{.857} & \textbf{.934} \\
        SRS & 2023 & W & .042 & .817 & .905 & .027 & .899 & .95 & .036 & .84 & .909 \\
		\hline
		TSD & 2023 & U & .047 & .807 & .903 & .032 & .881 & .939 & .044 & .805 & .899 \\
		Ours$_{t.s.}$ & - & U & .041 & .809 & .907 & .024 & .886 & .948 & .057 & .789 & .883 \\
		Ours & - & U & \textbf{.038} & \textbf{.843} & \textbf{.924} & \textbf{.023} & \textbf{.904} & \textbf{.956} & \textbf{.041} & \textbf{.846} & \textbf{.918} \\
		\hline 
	\end{tabular}
	\vspace{-1mm}
	\caption{Quantitative comparison on RGB-T SOD benchmarks. }
	\label{tab:RGBTSOD}
	\vspace{-2mm}
\end{table*}
\begin{table*}
	\centering
	\small
	\setlength\tabcolsep{1.5mm}
	\begin{tabular}{l|c|c|ccc|ccc|ccc|ccc}%
		\hline 
		\multicolumn{3}{c|}{dataset} & \multicolumn{3}{c|}{DAVSOD} & \multicolumn{3}{c|}{DAVIS} & \multicolumn{3}{c|}{SegV2} & \multicolumn{3}{c}{FBMS} \\
		\hline 
		Method & Year & Sup. & $M\downarrow\:$ & $F_\beta\uparrow\:$ & $E_\xi\uparrow\:$ & $M\downarrow\:$ & $F_\beta\uparrow\:$ & $E_\xi\uparrow\:$ & $M\downarrow\:$ & $F_\beta\uparrow\:$ & $E_\xi\uparrow\:$ & $M\downarrow\:$ & $F_\beta\uparrow\:$ & $E_\xi\uparrow\:$ \\
		\hline
        STVS & 2021 & F & .080 & .563 & .764 & .022 & .812 & .940 & .016 & .835 & .950 & .042 & .821 & .903 \\
		WSVSOD & 2021 & W & .103 & .492 & .710 & .036 & .731 & .900 & .031 & .711 & .909 & .084 & .736 & .840 \\
		\hline
		TSD & 2023 & U & .085 & .547 & .762 & .037 & .756 & .908 & .021 & .808 & .927 & .060 & .795 & .876 \\
		Ours & - & U & .092 & .572 & .754 & .041 & .764 & .897 & \textbf{.018} & \textbf{.842} & .92 & .052 & \textbf{.822} & .891 \\
		Ours$_{f}$ & - & U & \textbf{.084} & \textbf{.576} & \textbf{.764} & \textbf{.030} & \textbf{.793} & \textbf{.917} & .019 & .83 & \textbf{.936} & \textbf{.051} & .82 & \textbf{.896} \\
		\hline 
	\end{tabular}
	\vspace{-1mm}
	\caption{Quantitative comparison on video SOD benchmarks. }
	\label{tab:VSOD}
	\vspace{-2mm}
\end{table*}
\begin{table}
	\centering
	\small
	\setlength\tabcolsep{1mm}
	\begin{tabular}{l|c|c|ccc|ccc}%
		\hline 
		\multicolumn{3}{c|}{dataset} & \multicolumn{3}{c|}{ORSSD} & \multicolumn{3}{c}{EORSSD} \\
		\hline 
		Method & Year & Sup. & $M\downarrow\:$ & $F_\beta\uparrow\:$ & $E_\xi\uparrow\:$ & $M\downarrow\:$ & $F_\beta\uparrow\:$ & $E_\xi\uparrow\:$ \\
		\hline
		LVNet & 2019 & F & .021 & .751 & .92 & .015 & .628 & .845 \\
		MJRB & 2022 & F & \textbf{.016} & \textbf{.802} & \textbf{.933} & \textbf{.010} & \textbf{.707} & \textbf{.890} \\
		\hline
		Ours & - & U & .057 & .669 & .83 & \textbf{.053} & .545 & .755 \\
		Ours$_{f}$ & - & U & \textbf{.053} & \textbf{.726} & \textbf{.874} & .064 & \textbf{.625} & \textbf{.808} \\
		\hline 
	\end{tabular}
 	\vspace{-1mm}
	\caption{Quantitative comparison on RSI SOD benchmarks. }
	\label{tab:RSISOD}
	\vspace{-3mm}
\end{table}
\subsubsection{Results on RGB-D and RGB-T SOD}
Table \ref{tab:RGBDSOD} and \ref{tab:RGBTSOD} present a comparison between the proposed method 
and recent methods on RGB-D and RGB-T benchmarks, respectively. 
For a fair comparison, we also train our method using task-specified data, denoted as ``Ours$_{t.s.}$". 
VST~\cite{VST}, CCFE\cite{liao2022cross}, DSU~\cite{ji2022promoting}, TSD, MIDD~\cite{9454273} and SRS~\cite{liu2023scribble} are employed for comparison. 
Our method has achieved state-of-the-art performance on both RGB-D and RGB-T SOD. 
Moreover, our proposed approach, referred to as ``Ours", exhibits a substantial performance improvement compared to ``Ours$_{t.s.}$". 
We attribute this improvement to the limited size of the training datasets for these specific tasks. 
In contrast, ``Ours" was trained on a diverse range of datasets encompassing RGB, RGB-D, and RGB-T SOD, 
effectively utilizing shared common knowledge across different SOD tasks. 

\subsubsection{Results on video SOD and RSI SOD}
Table \ref{tab:VSOD}, \ref{tab:RSISOD} present a comparison between the proposed method and recent methods on video SOD and RSI SOD benchmarks, respectively. 
STVS~\cite{chen2021exploring}, WSVSOD~\cite{zhao2021weakly}, LVNet~\cite{li2019nested}, MJRB~\cite{9511336} and TSD are employed for comparison. 
We consider video SOD and RSI SOD as two types of target transfer tasks. 
Thus, in the table, ``Ours" represents zero-shot transfer results, 
while ``Ours$_{f}$" refers to the outcomes obtained by fine-tuning the transferred model on the target task using our proposed knowledge transfer approach. 
Note that the transfer for video SOD and RSI SOD is conducted separately. 
Our model exhibits excellent adaptability to the target task following fine-tuning, and exhibits remarkable performance. 
\subsection{Ablation study}
\begin{table}
	\centering
	\small
	\setlength\tabcolsep{0mm}
	\begin{tabular}{l|cc|cc|cc|cc|cc}%
		\hline 
		\multirow{2}{*}{Method} & \multicolumn{2}{c|}{RGB} & \multicolumn{2}{c|}{RGB-D} & \multicolumn{2}{c|}{RGB-T} & \multicolumn{2}{c|}{video} & \multicolumn{2}{c}{RSI} \\
  		\hhline{~|----------}
		 & $M\downarrow\:$ & $F_\beta\uparrow\:$ & $M\downarrow\:$ & $F_\beta\uparrow\:$ & $M\downarrow\:$ & $F_\beta\uparrow\:$ & $M\downarrow\:$ & $F_\beta\uparrow\:$ & $M\downarrow\:$ & $F_\beta\uparrow\:$  \\
		\hline
		Ours$_{t.s.}$ & \textbf{.033} & \textbf{.928} & .052 & .854 & \textbf{.019} & .949 & - & - & - & -\\
		Ours & \textbf{.033} & .927 & \textbf{.047} & \textbf{.87} & .020 & \textbf{.953} & \textbf{.068} & .696 & .074 & .634 \\
		Ours$_{f}$ & - & - & - & - & - & - & .070 & \textbf{.698} & \textbf{.051} & \textbf{.743} \\
		\hline 
	\end{tabular}
	\caption{Evaluation on Pseudo-label Quality. }
	\label{tab:LQ}
	\vspace{-3mm}
\end{table}
\begin{table}
	\centering
	\small
	\setlength\tabcolsep{1mm}
	\begin{tabular}{c|c|c|cc|cc|cc}%
		\hline 
	 	\multicolumn{3}{c|}{Refine Settings} & \multicolumn{2}{c|}{RGB} & \multicolumn{2}{c|}{RGB-D} & \multicolumn{2}{c}{RGB-T} \\
  		\hline
		$G_{\text{res}}$ & $R_{\text{pri}}$ & $R_{\text{post}}$ & $M\downarrow\:$ & $F_\beta\uparrow\:$ & $M\downarrow\:$ & $F_\beta\uparrow\:$ & $M\downarrow\:$ & $F_\beta\uparrow\:$  \\
		\hline
		\ding{51} & \ding{55} & \ding{55} & .04 & .918 & .064 & .825 & .03 & .923 \\	
		\ding{51} & \ding{55} & \ding{51} & .034 & .925 & .048 & .868 & .022 & .951\\
		\ding{51} & \ding{51} & \ding{51} & \textbf{.033} & \textbf{.927} & \textbf{.047} & \textbf{.87} & \textbf{.020} & \textbf{.953} \\
		\hline 
	\end{tabular}
	\caption{Evaluation on Self-rectify Pseudo-label Refinement. }
	\label{tab:SPR}
	\vspace{-3mm}
\end{table}
\subsubsection{Evaluation on Pseudo-label Quality}
We assessed the quality of the pseudo-labels generated by models trained on different datasets. 
As previously mentioned, ``Ours$_{t.s.}$" denotes the model trained using task-specific data, 
whereas ``Ours$_{f}$" refers to the model transferred to the target task. 
The results are presented in Table \ref{tab:LQ}. 
In comparison to ``Ours$_{t.s.}$", ``Ours" exhibits slightly inferior performance on the RGB training set, 
but displays a notable improvement on the RGB-D and RGB-T training sets, 
which possess a comparatively limited amount of training data. 
This indicates that a larger training dataset yields superior model performance and enhanced generalization ability. 
Furthermore, ``Ours$_{f}$" shows a slight improvement in video SOD, whereas it exhibits a substantial enhancement in RSI SOD. 
This indicates that video SOD and NSI SOD share more common knowledge, while RSI SOD requires greater fine-tuning and adaptation. 
More analysis on the adaptation to target tasks is presented in supplementary materials. 
\subsubsection{Evaluation on SPR}
We evaluated the influence of various rectifications on the pseudo-labels, 
and the results are presented in Table~\ref{tab:SPR}. 
The posterior rectification $R_{\text{post}}$ effectively corrects the erroneous predictions present in pseudo-labels, 
while the prior rectification $R_{\text{pri}}$ adequately compensates for the lack of local details in pseudo-labels. 
Through the combination of posterior correction and prior correction, 
the proposed SPR gradually enhances the quality of pseudo-labels, thereby improving the model's performance.
\subsubsection{Evaluation on Supervision Strategy}
\begin{table}
	\centering
	\small
	\setlength\tabcolsep{1mm}
	\begin{tabular}{l|cc|cc|cc}%
		\hline 
	 	\multirow{2}{*}{Loss Settings} & \multicolumn{2}{c|}{RGB} & \multicolumn{2}{c|}{DUTS-TE} & \multicolumn{2}{c}{NLPR} \\
  		\hhline{~|------}
		 & $M\downarrow\:$ & $F_\beta\uparrow\:$ & $M\downarrow\:$ & $F_\beta\uparrow\:$ & $M\downarrow\:$ & $F_\beta\uparrow\:$  \\
		\hline
		w/o PCL-SD & \textbf{.044} & .895 & .077 & .7 & .05 & .757 \\
		w/ PCL-SD & \textbf{.044} & \textbf{.896} & \textbf{.074} & \textbf{.713} & \textbf{.047} & \textbf{.77} \\
		\hline
		$\mathcal{L}_{bce}$ & .034 & .924 & .050 & .784 & .033 & .84 \\
		$\mathcal{L}_{iou}$ & .034 & .926 & .049 & .806 & .029 & .866 \\
		$\mathcal{L}_{iou}$+$\mathcal{L}_{bce}$ & \textbf{.033} & \textbf{.928} & .049 & .799 & .032 & .851 \\
		$\mathcal{L}_{iou}$+$\mathcal{L}_{ms}$ & \textbf{.033} & .927 & \textbf{.047} & \textbf{.816} & \textbf{.028} & \textbf{.871} \\
		\hline 
		 
	\end{tabular}
	\caption{Evaluation on Supervision Strategy. ``RGB" denotes the training set of RGB SOD. }
	\label{tab:loss}
	\vspace{-3mm}
\end{table}
We evaluated the effectiveness of the proposed supervision strategy, as shown in Table~\ref{tab:loss}. 
We treat all samples as easy samples to examine the effectiveness of PCL-SD. 
Upon applying PCL-SD, the model exhibits a slight improvement on the training set. 
Nonetheless, an impressive enhancement in performance can be observed on the test set. 
This improvement substantiates the model's heightened generalization capability. 
More experiments on the PCL-SD can be found in supplementary materials. 
Additionally, we explored the training of the saliency detector using different loss functions. 
The results indicate that the commonly employed binary cross-entropy (bce) in supervised SOD did not lead to effective performance enhancement. 
We hypothesize that this ineffectiveness may be attributed to the errors and interference stemming from incorrect predictions in pseudo-labels. 
In contrast, the self-supervised loss $\mathcal{L}_{ms}$ delivered a noteworthy improvement.

\section{Conclusion}
In this paper, we propose a two-stage unified unsupervised SOD framework for generic SOD tasks, with knowledge transfer as the foundation. 
Specifically, we introduce two innovative mechanisms: Progressive Curriculum Learning-based Saliency Distilling (PCL-SD) 
and Self-rectify Pseudo-label Refinement (SPR), 
which aim to extract saliency cues and optimize pseudo-labels. 
Additionally, we present a universal fine-tuning method to transfer the acquired saliency knowledge to generic SOD tasks. 
Extensive experiments on five representative SOD tasks validate the effectiveness and feasibility of our proposed method.

\section*{Acknowledgements}
This work was partially supported by the National Natural Science Foundation of China (No. 62272227 \& No. 62276129), and the Natural Science Foundation of Jiangsu Province (No. BK20220890).

\small
\bibliographystyle{named}
\bibliography{ijcai24}

\section{Supplement Materials}
\subsection{Motivation}
Unsupervised salient object detection (USOD) has emerged as a solution to address the reliance on large-scale annotated data, enabling models to train on more extensive datasets. The annotation-free nature of unsupervised SOD facilitates stronger generalization performance. However, in the case of certain fundamental SOD datasets (e.g., MSRA~\cite{4270072}, DUTS~\cite{DUTS}, VT5000~\cite{tu2022rgbt}), substantial annotated data already exists, and existing supervised methods have achieved good performance. When it comes to basic Natural Still Image (NSI) SOD, such as RGB, RGB-D, and RGB-T tasks, unsupervised methods can only avoid overfitting on annotations but still exhibit significant performance gaps compared to supervised methods. 
However, for more advanced SOD tasks, such as video SOD and remote sense image (RSI) SOD, the challenges lie not only in annotation difficulties but also in high data collection thresholds or costs. Limited datasets in these scenarios make it prone to overfitting and overconfidence during training. Additionally, the limited amount of data makes training a model from scratch without supervision challenging to achieve desirable performance. Currently, no unsupervised methods exist for RSI-SOD, and research on unsupervised SOD in these tasks has been largely neglected. 

We posit that there is a correlation between different SOD tasks, implying the presence of shared common knowledge. Existing studies have attempted to develop a universal approach by incorporating multiple SOD tasks into a unified framework. Specifically, ~\cite{wang2023unifiedmodal} attempted prompt tuning to address the inconsistency between modalities. However, these investigations are largely confined to the NSI domain. As we broaden our perspective to encompass a wider range of SOD domains, the shared knowledge between tasks becomes more limited, and the differences become the primary influencing factors. In this paper, we introduce video SOD and RSI SOD tasks to establish the notion of generic SOD, which can also include other specific tasks like hyperspectral salient object detection, underwater salient object detection, and so on. Therefore, training a unified model by combining data from all tasks to address the generic SOD task is currently impractical. 

Despite the increase in differences, we firmly hold the belief that common saliency knowledge exists. Specifically, we contend that video SOD and RSI SOD tasks can learn valuable knowledge from the NSI domain. Unsupervised approaches, unlike supervised methods, are not hindered by manual annotations, enabling them to focus on this shared common knowledge and exhibit better generalization and transferability. Previous research~\cite{zhou2023texture} has further demonstrated that unsupervised methods achieve superior zero-shot transfer results compared to existing supervised methods in scenarios such as chest X-ray images. Based on the aforementioned analysis, we assert that the exploration of unsupervised SOD methods for generic SOD tasks is both valuable and meaningful. It fosters research in data-scarce SOD tasks and propels SOD advancements in real-world scenarios. 

\subsection{Details of the proposed framework}
\begin{figure*}
	\centering 
	\includegraphics[scale=0.85]{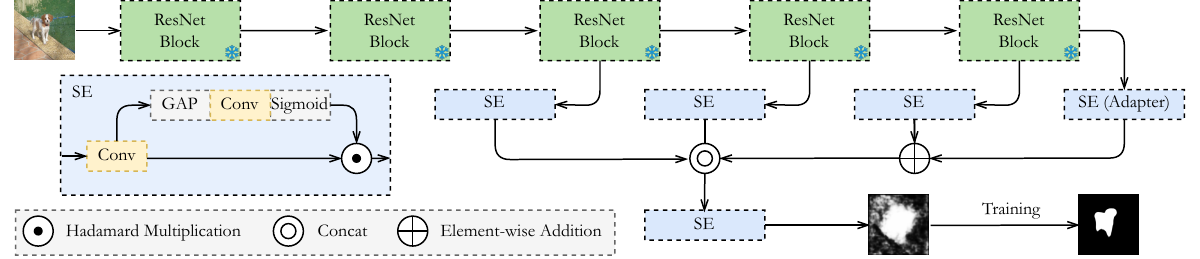}
	\caption{Architecture of the saliency cue extractor. }
	\label{fig:sce}
\end{figure*}
\begin{figure*}
	\centering 
	\includegraphics[scale=0.85]{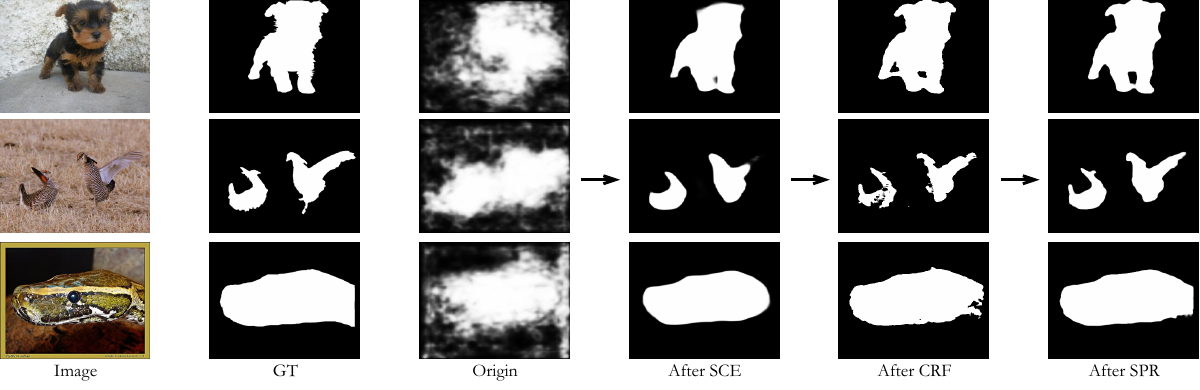}
    \caption{The optimization process of pseudo labels. }
	\label{fig:pseudo}
\end{figure*}
Our proposed framework employs a saliency cue extractor (SCE) and a saliency detector (SD). The architecture of the SCE is illustrated in Figure~\ref{fig:sce}. We utilize A2S~\cite{zhou2023a2s1} as the underlying model and incorporate an adapter component for transfer learning. 
The SCE consists of ResNet as the backbone, along with several SE~\cite{hu2018senet} modules. Its objective is to obtain a multi-scale activation map from different layers of the backbone. The structure of SCE is similar to existing end-to-end SOD models, but deliberately kept simple to control network depth and obtain activation map outputs closer to the backbone. Whether training from scratch or fine-tuning, we maintain weight freezing of the backbone. This is because unsupervised training is susceptible to instability, and freezing the weights helps mitigate the risk of model degradation. 
During fine-tuning for transfer learning to other tasks, we assume that deep features are more task-specific, while shallow features are generally more general. Based on this assumption, we apply adapter tuning only to the deepest layer's features. Specifically, we add an additional SE module and use residual connections to add its output to the original network. Only the parameters of this module are involved in gradient backpropagation, while the remaining parameters are kept frozen. 

For the saliency detector, we utilize MIDD~\cite{9454273} directly for experimentation without any modifications. Our framework allows for the integration of existing models that are designed for dual-stream input tasks, such as RGB-D and RGB-T, including VST~\cite{VST} and CAVER~\cite{CAVER-TIP2023}. In fact, models originally developed for RGB SOD tasks using only RGB input can also be employed. However, in such cases, we are unable to obtain effective saliency references from the modal input. It is worth mentioning that models that perform better under supervised training may not necessarily exhibit the same superiority under our unsupervised training. This discrepancy can be attributed to two factors: 1. We use pseudo labels instead of ground truth for training, and pseudo labels inherently contain some level of noise. Therefore, the model's ability to resist noise interference will affect its performance. 2. Our Self-rectify Pseudo-label Refinement (SPR) mechanism involves refining pseudo labels based on the model's outputs. If the model's outputs not only fail to suppress the noise in pseudo labels but also introduce new erroneous predictions, then the model's performance will deteriorate with further training. 
\begin{figure}
	\centering 
	\includegraphics[scale=0.85]{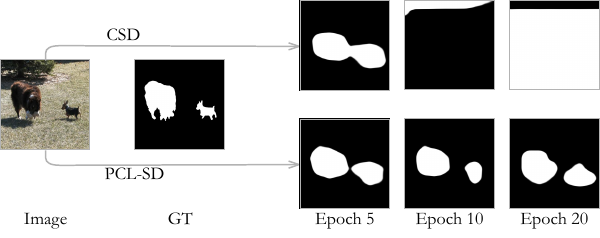}
	\vspace{-6mm}
    \caption{The comparison between proposed PCL-SD and CSD. }
	\label{fig:pcl}
\end{figure}
\subsection{Form high-quality pseudo-labels from scratch}
In this section, we present the process of generating pseudo labels from scratch. As stated in the main text, the initial stage of our approach involves training a saliency cue extractor (SCE) to extract saliency cues from a pre-trained deep network. This procedure gradually transforms the output of the SCE from cluttered and unordered activation maps into saliency cues that accurately reflect the positions of salient objects, as illustrated in Figure~\ref{fig:pseudo}. 

To obtain activation maps that closely resemble those of the deep network, we must maintain a simple design for the SCE, resulting in relatively poor structural quality of the extracted saliency cues. Therefore, we employ Conditional Random Fields (CRF) to enhance these saliency cues and obtain initial pseudo labels. Subsequently, we train an additional saliency detector (SD) using these pseudo labels. CRF serves as a common post-processing technique that utilizes prior information about the image to compensate for the suboptimal structural quality of the saliency cues. Nevertheless, due to the absence of guidance from high-level semantic information and the influence of complex environments, the initial pseudo-labels obtained through this process often contain noise. Therefore, during the training of SD, we leverage the high-level semantic information present in the output of SD to rectify erroneous predictions within the pseudo labels. 

Here, we would like to discuss further the prior and posterior rectifications incorporated in the proposed Self-rectify Pseudo-label Refinement. The posterior rectification pertains to the output of the saliency detector, while the prior rectification can refer to either CRF or the real-time pixel refiner adopted in our methodology. We apply CRF only once to acquire the initial pseudo labels, and all subsequent prior corrections are performed using the pixel refiner. This choice is justified by the time-consuming nature of CRF, as its processing duration for an image considerably exceeds the time required for the model to make predictions on the same image. Following the SPR process, the quality of the pseudo labels is further improved, as demonstrated in Figure~\ref{fig:pseudo}. 

\subsection{Curriculum Learning and SOD}

\begin{figure*}
	\centering 
	\includegraphics[scale=0.85]{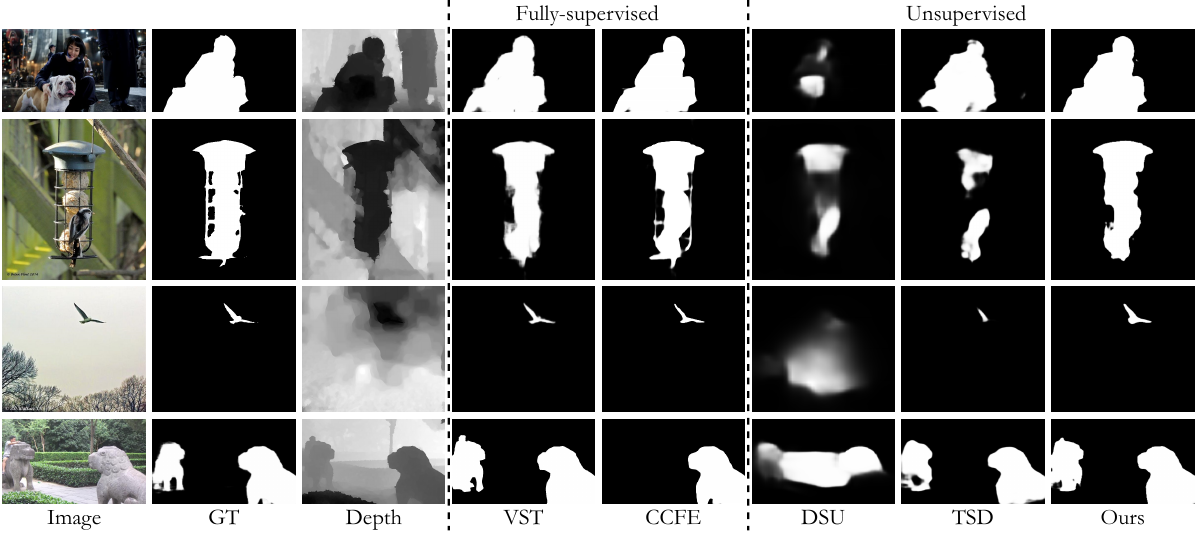}
	\caption{Visual comparison between the proposed method and the other state-of-the-art SOD methods on RGB-D SOD datasets. }
	\label{fig:rgbd}
\end{figure*}
\begin{figure*}
	\centering 
	\includegraphics[scale=0.85]{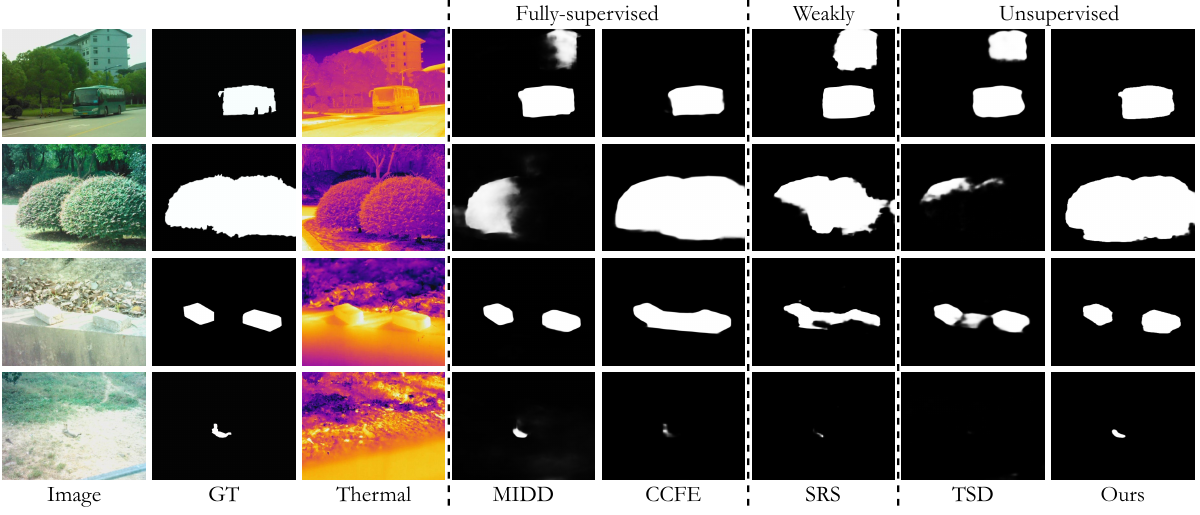}
	\caption{Visual comparison between the proposed method and the other state-of-the-art SOD methods on RGB-T SOD datasets. }
	\label{fig:rgbt}
\end{figure*}
\begin{figure*}
	\centering 
	\includegraphics[scale=0.85]{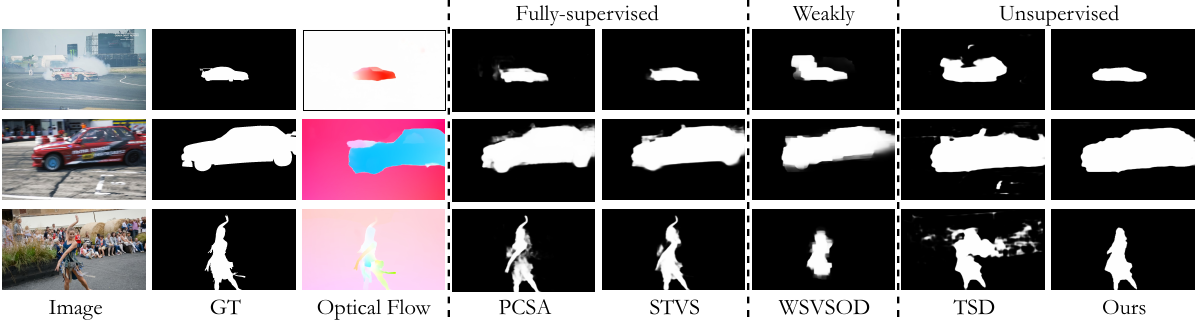}
	\caption{Visual comparison between the proposed method and the other state-of-the-art SOD methods on video SOD datasets. }
	\label{fig:video}
\end{figure*}
To the best of our knowledge, we are the first to introduce the concept of curriculum learning into the field of salient object detection (SOD). Curriculum learning, originating from NLP tasks, centers around the idea of learning simple knowledge before tackling more challenging concepts. Designing a curriculum learning method involves focusing on three crucial aspects: (1) defining hard samples, (2) gradually incorporating hard samples, and (3) dealing with hard samples. In the main text, we discuss the first two points when presenting our proposed PCL-SD mechanism. For the third aspect, there are two strategies: a hard partition strategy, which strictly removes all hard samples by excluding them from backpropagation, and a soft partition strategy, which initially assigns a lower learning weight to the hard samples and gradually increases it. The latter still suffers from the problem of error accumulation caused by hard samples, leading to unstable training. Therefore, we adopt the former strategy, which rigidly removes all hard samples. 

We conducted comparative experiments between our proposed PCL-SD and CSD~\cite{zhou2023texture}. Surprisingly, we discovered that models trained with CSD are at risk of pattern collapse. As shown in Figure~\ref{fig:pcl}, at epoch 10, the model's outputs became meaningless and unrelated to the input images, consisting of large areas of pure white or black. We cannot solely attribute the phenomenon of pattern collapse to the interference of hard samples. However, when applying our proposed PCL-SD, the model training becomes more stable. Thus, we can conclude that PCL-SD effectively mitigates the problem of error accumulation caused by hard samples in the early stages of training, resulting in a more stable and robust training process. 

In addition, we have also attempted to apply this hard sample handling strategy to supervised training. In this case, the determination of whether a specific point is a hard sample is based on its corresponding loss value, with a higher loss indicating increased difficulty. Regrettably, this approach did not yield any additional improvements. We suppose that two factors may contribute to this outcome: 1) the removal of hard samples impedes the model's ability to perceive intricate regions, and 2) the removal of hard samples introduces a disconnection between samples. To address the first factor, when designing PCL-SD, we only restrict the involvement of hard samples in gradient backpropagation during the initial few epochs of training, thereby avoiding the model's inability to fully learn from the knowledge contained within hard samples. As for the second factor, when calculating the loss using $L_{sd}$, each sample is treated as independent thereby no disconnection occurs. Overall, we believe that if these two factors can be avoided, curriculum learning still has good potential in SOD field. 

\subsection{More comparison results}

\begin{figure}
	\centering 
	\includegraphics[scale=0.85]{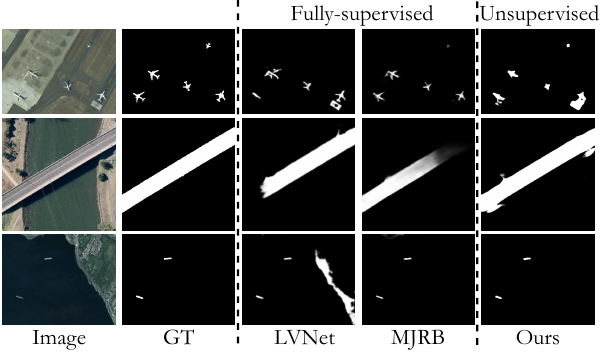}
	\caption{Visual comparison between the proposed method and the other state-of-the-art SOD methods on RSI SOD datasets. }
	\label{fig:rsi}
\end{figure}
We present visual comparisons of our proposed unsupervised method with other state-of-the-art methods on four tasks: RGB-D SOD (Figure~\ref{fig:rgbd}), RGB-T SOD (Figure~\ref{fig:rgbt}), video SOD (Figure~\ref{fig:video}), and Remote Sensing Image (RSI) SOD (Figure~\ref{fig:rsi}). 
For RGB-D SOD, the compared methods include VST~\cite{VST}, CCFE~\cite{liao2022cross}, DSU~\cite{ji2022promoting}, and TSD~\cite{zhou2023texture}. 
For RGB-T SOD, the compared methods include MIDD~\cite{9454273}, CCFE, SRS~\cite{liu2023scribble}, and TSD. 
For video SOD, the compared methods include PCSA~\cite{gu2020PCSA}, STVS~\cite{chen2021exploring}, WSVSOD~\cite{zhao2021weakly}, and TSD. 
For RSI SOD, the compared methods include LVNet~\cite{li2019nested} and MJRB~\cite{9511336}. The visual comparisons of RGB-D and RGB-T tasks demonstrate that our proposed method not only surpasses existing unsupervised methods but also achieves competitive performance compared to supervised methods. Moreover, for video SOD and RSI SOD tasks, we effectively leverage shared saliency knowledge transferred from the Nature Still Image (NSI) SOD tasks, resulting in impressive performance without any annotation data. These visual results confirm the outstanding performance of our proposed USOD framework and demonstrate the effectiveness of the proposed knowledge transfer approach. 

\subsection{Limitations \& Future works}

In this paper, we propose a unified unsupervised framework for salient object detection (SOD) based on knowledge transfer. While our method has achieved remarkable results, there are still areas for further improvement. We highlight three key directions for future research. 

\subsubsection{From Modality-Agnostic to Modality-Informed}
During the training of our saliency detector on Nature Still Image (NSI) data, the model treats different modalities, such as depth, thermal, and optical flow, in a modality-agnostic manner. This approach does not specifically differentiate the input modality, treating it as generic reference information. While this enhances the model's generalization and robustness, it may limit the model's ability to exploit saliency information specific to each modality. Hence, transitioning from a modality-agnostic to a modality-informed approach is essential to facilitate better learning of shared saliency knowledge across different SOD tasks. 

\subsubsection{From Two-Stage to One-Stage}
In this work, we adopt a two-stage framework, similar to previous unsupervised SOD (USOD) methods. This involves obtaining saliency clues or pseudo labels in the first stage and training a saliency detector using these labels in the second stage. Although this two-stage training approach helps mitigate the instability associated with unsupervised methods, it may lead to potential disconnections. Mislocalized salient objects in the first stage are challenging to correct in the second stage. Therefore, exploring a one-stage model that addresses these issues is a worthwhile consideration. 
\begin{figure}
	\centering 
	\includegraphics[scale=0.85]{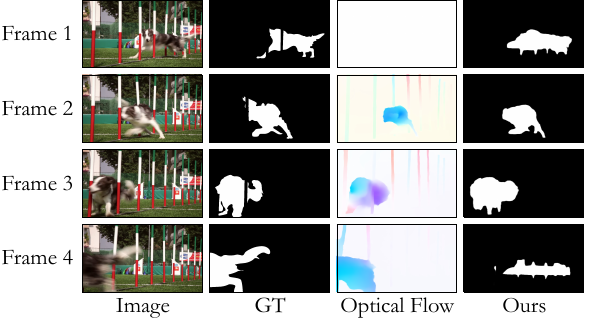}
	\caption{Our approach considers each frame within the video as an independent two-dimensional image for saliency detection. It effectively identifies salient objects in Frames 1, 2, and 3, yet encounters troubles in detecting salient objects in Frame 4. }
	\label{fig:video_}
\end{figure}
\subsubsection{Deeper and More Targeted Migration}
In our work, we mainly focus on migrating to video SOD and RSI SOD. However, we treated both tasks as non-NSI SOD tasks without incorporating more targeted adaptation measures. For instance, in video SOD, we did not directly input video data into the model but treated each frame as a separate two-dimensional image. Although this approach better utilizes saliency knowledge transferred from NSI SOD, it leads to issues depicted in Figure~\ref{fig:video_}, where saliency is assessed based on the object's presence throughout the entire video, even if it is not salient in a specific frame. In future research, we plan to address these challenges and explore more tailored strategies specifically designed for these types of tasks. 

\end{document}